\documentclass[letterpaper]{article} 
\usepackage{aaai23}  
\usepackage{times}  
\usepackage{helvet}  
\usepackage{courier}  
\usepackage[hyphens]{url}  
\usepackage{graphicx} 
\usepackage{natbib}  
\usepackage{caption} 
\usepackage{algorithm}
\usepackage{algorithmic}
\usepackage{arydshln}
\usepackage{multirow}
\usepackage{mathtools}
\usepackage{subcaption}
\usepackage{enumitem}
\usepackage{newfloat}
\usepackage{listings}
\DeclareCaptionStyle{ruled}{labelfont=normalfont,labelsep=colon,strut=off} 
\floatstyle{ruled}
\newfloat{listing}{tb}{lst}{}
\floatname{listing}{Listing}
\newcommand{\fscore}{F1-score}

\title{\textsc{SumREN}: Summarizing Reported Speech about Events in News}
\author{
    Revanth Gangi Reddy\textsuperscript{\rm 1}\thanks{Work primarily done during an internship at Amazon Alexa.}, Heba Elfardy\textsuperscript{\rm 2}, Hou Pong Chan\textsuperscript{\rm 3}, Kevin Small\textsuperscript{\rm 2}, Heng Ji\textsuperscript{\rm 2}
}
\affiliations{
\textsuperscript{\rm 1}University of Illinois Urbana-Champaign \hspace{1em} \textsuperscript{\rm 2}Amazon Alexa\hspace{1em}
\textsuperscript{\rm 3}University of Macau\hspace{1em} \\
\texttt{revanth3@illinois.edu},\hspace{0.5em}\texttt{\{helfardy,smakevin,jihj\}@amazon.com},\hspace{0.5em}\texttt{hpchan@um.edu.mo}
}

\usepackage{comment}
\usepackage[dvipsnames]{xcolor}
\definecolor{demphcolor}{gray}{.45}
\newcommand{\demph}[1]{\textcolor{demphcolor}{#1}}
\usepackage{xparse}
\usepackage{bibentry}

\begin{document}

\maketitle

\begin{abstract}

A primary objective of news articles is to establish the factual record for an event, frequently achieved by conveying both the details of the specified event (i.e., the 5 Ws; Who, What, Where, When and Why regarding the event) and how people reacted to it (i.e., reported statements). However, existing work on news summarization almost exclusively focuses on the event details. 
In this work, we propose the novel task of summarizing the reactions of different speakers, as expressed by their reported statements, to a given event. To this end, we create a new multi-document summarization benchmark, \textsc{SumREN}, comprising 745 summaries of reported statements from various public figures obtained from 633 news articles discussing 132 events.\footnote{Code and data will be available at: \url{https://github.com/amazon-science/SumREN}} We propose an automatic silver-training data generation approach for our task, which helps smaller models like BART achieve GPT-3 level performance on this task. Finally, we introduce a pipeline-based framework for summarizing reported speech, which we empirically show to generate summaries that are more abstractive and factual than baseline query-focused summarization approaches. 
\end{abstract}

\section{Introduction}



\begin{table}[!htb]
\renewcommand{\arraystretch}{0.9}
\scriptsize
    \centering
    \begin{tabular}{|p{32em}|}
                 \hline
                 \multicolumn{1}{|c|}{\textbf{Event: \textit{Power Outage in Texas} \hspace{4em}Speaker: \textit{Nateghi}}} \\
                 \hline
                 \hline
                 \multicolumn{1}{|c|}{\textbf{Reported Statements}}\\
                 \hline
                 An issue facing all power grid operators, Nateghi of Purdue said, is adequately preparing for changes in climate. \\
                 \hline
                 They're also not taking into account inter-dependencies in the system: You need water to generate electricity, and you need electricity to transport water, and so on and so forth, Nateghi said. \\
                 \hline
                 And when the system is really stressed from an extreme event like it is in Texas, then we're seeing natural gas shortages which exacerbate the whole impact, she said. \\
                 \hline
                 Nateghi, who researches sustainability and resilience of infrastructure, said other solutions such as upgraded equipment and infrastructure may not be as cost-effective, but are still crucial.\\
                 \hline
                 ``If we continue down the paradigm of what we've done before we are going to see more extremes,'' Nateghi said. ``These stories are going to just keep playing, and perhaps even more frequently.'' \\
                 \hline
                 \hline
                 \textbf{Summary:} \textit{Nateghi said that interdependencies in the system are not being considered, and the problem of gas shortages could be seen in the power outage in Texas. Solutions such as upgraded equipment and infrastructure maybe less cost-effective but crucial. She also said that power grid operators needed to make changes before extreme situations became more frequent.}\\
                 \hline
\end{tabular}
    \caption{An example from \textsc{SumREN} showing reported statements from the speaker ``Nateghi'' about the ``Power outage in Texas'' along with the corresponding summary.}
    \label{tab:intro_ex}
    \vspace{-1.0em}
\end{table}
In news, attribution occurs when the journalist reports the statements of a third party either by directly quoting them (i.e., direct quotation) or paraphrasing what they said (i.e. indirect quotation). Reported speech serves as a central resource for tracking public figures' stance, opinions, and worldviews, making it of general interest to news readers. For example, readers are likely to be interested in knowing President Biden's view on the 2022 Ukraine crisis or the latest guidance from the Center for Disease Control and Prevention regarding a new COVID-19 variant. In addition, reported statements cover a significant portion of the information presented in news articles -- as part of our annotation exercise (described later in section \ref{sec:benchmark-construction}), we found that 45\% of the overall article content corresponds to reported statements. 
However, current news summarization datasets such as CNN-DM \cite{hermann2015teaching}, Multi-News \cite{fabbri2019multi}, and Timeline$_{100}$ \cite{li2021timeline} largely disregard summarizing these reported statements. 

To bridge this gap, we introduce the new task of \textbf{Sum}marizing \textbf{R}eported speech about \textbf{E}vents in \textbf{N}ews and create a new benchmark, \textsc{\textbf{SumREN}}, for this task. Formally, given a set of news articles related to a specified event, the task is to summarize the statements made by a given speaker about this event (e.g., ``What did Chuck Schumer say about passing the Inflation Reduction Act of 2022?"). The aim of the task is to provide news readers with the reactions of various public figures towards different events. Table \ref{tab:intro_ex} shows an example from \textsc{SumREN}, along with the reported statements and the corresponding reference summary. 

Summarizing reported speech in news brings a set of unique challenges. 
As opposed to traditional news summarization datasets where the most salient information about the event is normally discussed in the first few sentences of a given article, generally referred to as ``\textit{lead bias}'' \cite{jung2019earlier, zhu2021leveraging}, reported speech from the same speaker can be scattered across the entire article. Statements can be split across multiple sentences (i.e., ``running quotations'') and speakers are often referred to by their nominal and pronominal mentions, requiring modelling of long-term dependencies and reliable co-reference resolution.   
Additionally, generating concise summaries from a set of reported statements requires a higher level of abstraction. This is also verified empirically, as we find that summaries in \textsc{SumREN} are considerably more abstractive compared to existing news summarization datasets, as shown in Table \ref{tab:dataset_novel_ngrams}. 
Finally, factual consistency is paramount in reported speech summarization, as misquoting or misrepresenting statements from public figures can be particularly harmful.

To address the above challenges, we propose a pipeline-based approach for the task of summarizing reported speech in news articles. The pipeline involves first identifying individual statements and corresponding local speaker mentions, then resolving speaker mentions globally using co-reference resolution to group statements from the same speaker together, and finally summarizing the extracted reported statements by the given speaker. 
We hypothesize that, in a pipeline-based framework, having an explicit extractive component that can identify relevant context helps the summarization model better attend to the key information from the given articles.

In addition, we introduce a cost-effective approach to generating training data for the reported speech summarization task. Specifically, we leverage large-scale pre-trained language models, such as GPT-3~\cite{brown2020language}, to generate silver-standard summaries for statements obtained from automatic reported speech extraction systems. This follows recent work that uses large language models to create training data~\cite{schick2021generating}, although previously explored for discriminative tasks such as Natural Language Inference. We show that training with such silver-standard data can help smaller language models, such as BART \cite{lewis2020bart} achieve GPT-3-level performance on this task. \\ \\
To summarize, the contributions of this work include:
\begin{itemize}[noitemsep]
    \item introducing a new challenging task of summarizing reported speech about events in news and releasing the first multi-document summarization benchmark, \textsc{SumREN}, for the task. \textsc{SumREN} contains 745 instances annotated over 633 news articles discussing 132 events, 
    \item empirically demonstrating that large-scale language models can be leveraged to create cost-efficient silver-standard training data for the reported speech summarization task,
    \item proposing a pipeline-based reported speech summarization framework and showing that it is capable of generating summaries that are considerably more abstractive than query-focused approaches, while also improving the factual consistency of the generated summaries with the source documents.
    
\end{itemize}

\begin{figure*}
\centering
\scriptsize
\renewcommand{\arraystretch}{1.2}
      \begin{minipage}[]{0.55\textwidth} 
             
             \begin{tabular}{|p{35.5em}|}
                 \hline
                 \multicolumn{1}{|c|}{\textbf{Step 1: Identifying salient spans in statements}} \\
                 \hline
                 While \textcolor{red}{Republicans look inward} at the aftermath of the Capitol Hill riots after President Trump's address Wednesday, \textcolor{red}{Democrats are adding to the division}, Fox News contributor Charles Hurt told ``Fox \& Friends.''\\
                 \hline
                 ``I get this \textcolor{red}{rush to want to blame everything on President Trump}. Everything that is going on right now has been in the making for years and decades, of which politicians on Capitol Hill have been a part," Hurt, Washington Times opinion editor, told co-host Brian Kilmeade. \\
                 \hline
                 He added: ``The \textcolor{red}{last thing they want to do is take stock of themselves} and try to figure out, `OK, what have I done to make this worse or to create this situation?''' \\
                 \hline
                 Within seconds of reconvening Wednesday night, \textcolor{red}{Democrats on Capitol Hill started ``accusing Republicans of treason and sedition,''} Hurt said. \\
                 \hline
                 ``\textcolor{red}{They get caught up in their own mob mentality, they're all trying to outdo one another} on Twitter to see who can make the most outrageous charge or make the most outrageous demand of the other side,'' Hurt said. \\
                 \hline
                 While this might be a \textcolor{red}{good time for soul-searching} for both parties, Hurt concluded, ``There is \textcolor{red}{no indication from Democrats} on Capitol Hill \textcolor{red}{that any one of them has any intention of doing that} and certainly not from Joe Biden or Kamala Harris.''\\
                 \hline
             \end{tabular}
     \end{minipage}
     \hfill
     \begin{minipage}[]{0.445\textwidth}
         \begin{tabular}{|p{30em}|}
             \hline
             \multicolumn{1}{|c|}{\textbf{Step 2: Grouping salient spans into sentences.}} \\
              \hline
              ``blame everything on President Trump'' + ``accusing the Republicans of treason and sedition'' \newline
              $\rightarrow$ \textcolor{blue}{\textit{``blame everything on President Trump and accuse the Republicans.''}} \\
              \hline
              ``Democrats are adding to the division'' + ``they get caught up in their own mob mentality, they're all trying to outdo one another'' \newline
              $\rightarrow$ \textcolor{blue}{\textit{``Democrats get caught up in trying to outdo one another and are adding to the division.''}} \\
              \hline
               ``last thing they want to do is take stock of themselves'' + ``this might be a good time for soul-searching'' + ``no indication from Democrats that any one of them has any intention of doing that'' \newline
              $\rightarrow$ \textcolor{blue}{\textit{``They don’t seem to have any intention of doing any soul-searching''}} \\
              \hline
         \end{tabular}
         \vspace{1.0em}
         
         \begin{tabular}{|p{30em}|}
             \hline
             \multicolumn{1}{|c|}{\textbf{Step 3: Combining sentences into a summary}} \\
             \hline
             \textit{Charles Hurt suggested that Democrats are rushing to \textcolor{blue}{blame everything on President Trump and accuse the Republicans}. He said that \textcolor{blue}{Democrats get caught up in trying to outdo one another and are adding to the division}. Finally, \textcolor{blue}{they don’t seem to have any intention of doing any soul-searching}.}\\
             \hline
         \end{tabular}
     \end{minipage}
     \caption{Walk-through example showing the process of annotating summaries given a set of reported statements. Salient spans within the statements are shown in red and sentences copied over from step 2 into the summary in step 3 are shown in blue.}
     \label{fig:summ_example}
\vspace{-0.5em} 
\end{figure*}

\section{Related Work}

Our work draws from multiple related research veins
as itemized in this section.
\paragraph{News Summarization:} Summarizing news articles has been extensively studied in existing literature with multiple existing datasets. Single document summarization datasets include CNN/Daily Mail~\cite{hermann2015teaching},  News Room corpus \cite{grusky2018newsroom} and the XSum dataset~\cite{narayan2018don}. \citet{fabbri2019multi} introduce a large-scale dataset, Multi-News, to extend news summarization to a multi-document setting.  Timeline summarization \cite{steen2019abstractive, li2021timeline} adds a temporal aspect to news summarization by generating a sequence of major news events with their key dates. 
Another line of work lies around news headline generation \cite{banko2000headline}, which involves generating representative headlines for a given news story, explored in both single- \cite{hayashi2018headline} and multi-document settings \cite{gu2020generating}. However, these datasets all largely focus on summarizing the details of events and neglect the reported speech related to these events.


\paragraph{Query-Focused Summarization:}
Query-focused summarization (QFS) aims to produce a summary that answers a specific query about the source document(s). Conceptually, reported speech summarization corresponds to the query, {\em ``What did X say about Y?"}.
Prior work builds large-scale QFS datasets by obtaining reference summaries by scraping them from the web or using pseudo-heuristics. For example,  WikiSum~\cite{DBLP:conf/iclr/wikisum18} and AQuaMuSe~\cite{DBLP:journals/corr/aquamuse2020} directly extract paragraphs from Wikipedia articles as reference summaries. On the other hand, manually annotated QFS datasets are small -- DUC 2006 and 2007 \cite{dang2005overview} contain up to only 50 examples. QMSum~\cite{zhong2021qmsum} focuses on summarizing meeting dialogue transcripts and is most similar to our work. However, QMSum transcripts contain a considerable amount of informal conversations and do not contain focused informative content like the reported statements in \textsc{SumREN}. 

Since QFS datasets usually come with only source-summary pairs, most prior work either use end-to-end approaches  \cite{vig2021exploring,DBLP:journals/tacl/LatenQuery22} or follow a two-step extract-then-abstract framework \cite{DBLP:conf/acl/XuL20, vig2021exploring}, with the extractor trained to identify text spans that are similar to the reference summary in terms of ROUGE scores. Conversely, \textsc{SumREN} additionally provides the corresponding relevant content, reported statements in this case, that was used to annotate the summaries. Thereby, our proposed pipeline-based approach can leverage this to build and evaluate an extractive component that is independent of the reference summary, while still ensuring the generated summary has high input fidelity in terms of factual consistency.
\paragraph{Attribution in News:}  Attribution has been well-studied with multiple available datasets. \citet{elson2010automatic, zhang2021directquote} study attribution of direct quotations along with their speakers. 
\citet{pareti2012database, pareti2013automatically} extend this notion by including indirect quotations and create the PARC3 corpus. 
More recently, PolNeAR \cite{newell2018attribution} was created to improve upon PARC3 by doubling the recall and improving inter-annotator agreement. However, all of these lines of work solely deal with identifying attribution and do not aggregate extracted statements from specific speakers to help with a downstream task. More direct uses of quotations in news include opinion mining \cite{balahur2009opinion} and sentiment analysis \cite{balahur2013sentiment}.  In contrast, our proposed task involves attribution to identify reported statements in news articles, which are then aggregated and summarized to convey the reactions to events in news. 


\section{SumREN Benchmark}

\label{sec:benchmark}
The SumREN benchmark aims to assist in the development and evaluation of models for the reported speech summarization task. 
In this section, we describe the task of reported speech summarization, the benchmark construction process, as well as present statistics of the constructed dataset.

Given a set of news articles about a specific event and the speaker name, the goal is to generate a succinct summary for the statements made by the speaker in the source content. 


\subsection{Benchmark Construction}

\label{sec:benchmark-construction}
The first step in our benchmark construction process involves collecting a news corpus discussing a large set of events. We split the news articles according to the discussed event and from each cluster of news articles, we then extract all reported statements along with the speakers of each of these statements. Finally, a summary is written for each group of statements by the same speaker.
\subsubsection{News Corpus Acquisition:} 
We first identified a list of 132 major news events between 2013-2021 that were mentioned in Wikipedia and other sources.
We then collected a list of news articles discussing these events and retained articles that are present in Common Crawl (CC) News.\footnote{For articles between 2013 and 2016, we relied on WayBack Machine since CC News is not available for these years.} We ended up with a total of 633 news articles corresponding to 132 major events.

\subsubsection{Reported Statements Annotation:} 
To annotate the reported statements and the speakers, we used Amazon Mechanical Turk and collected three annotations per HIT. The annotation tasks were restricted to annotators in English-speaking countries and who passed the custom qualification test for the corresponding task  -- reported statement span selection or speaker identification.\footnote{Please refer to appendix for detailed annotation guidelines.}
Overall, 12\% of the annotators that took the test were qualified. In addition, we blocked spammers that spent less than a specified number of seconds per task or that consistently provided low-quality annotations. 
For the reported statement span selection task, annotators were provided with a snippet from the news article and were asked to highlight the spans containing reported statements. Contiguous sentences with statements from the same speaker were considered to be parts of the same reported statement. After collecting the annotations, we grouped reported statements (and associated articles) by a specific speaker about each event.


\subsubsection{Summary Generation:}
 
For summary generation, we relied on expert annotators since it is a more challenging task and hence less suitable for MTurk. 
Two reference summaries produced by two different annotators were created for each cluster of reported statements. 
An abridged version of the annotation guidelines is presented below and a walk-through example of the annotation process is shown in Figure \ref{fig:summ_example}.
\begin{itemize}[noitemsep]
    \item \textbf{Step 1:} For each one of the given statements, identify the salient spans.
    \item \textbf{Step 2:} Group similar salient spans -- that discuss related aspects of the event -- together and combine these similar spans into a single sentence; \textit{using paraphrasing if needed}.
    \item \textbf{Step 3:} Combine these sentences into a summary.
\end{itemize}
\subsection{Statistics}
Our benchmark has 745 examples in total, with a train/dev/test split of 235/104/406 respectively. On average, the summaries have a length of 57 words and each summary comes from 5.3 reported statements. 57\% of the examples have a single source news article, with 26\% having 2 source articles and remaining 17\% having 3-5 source articles. The average combined source length is 2,065 words. Overall, the news corpus contains 633 articles with a total of 10,762 reported statements from 3,725 unique speakers. Further, we observe that the summaries in our benchmark are relatively more abstractive compared to existing summarization datasets. Table \ref{tab:dataset_novel_ngrams} shows the percentage of novel n-grams, with \textsc{SumREN} containing considerably higher novel tri-grams and 4-grams. To account for this relatively higher abstractiveness and also variance in generation, each example in our benchmark has two reference summaries.
\begin{table}[!htb]
\footnotesize
    \centering
    \def\arraystretch{1.2}
    \begin{tabular}{l|cccc}
    \hline
    \textbf{Datasets} & \textbf{unigram} & \textbf{bigram} & \textbf{trigram}  & \textbf{4-gram}\\
    \hline
    CNN-DM (S) & 17.0 & 53.9 & 72.0 & 80.3 \\
    NY Times (S) & 22.6 & 55.6 & 71.9 & 80.2 \\
    MultiNews (M) & 17.8 & 57.1 & 75.7 & 82.3 \\
    WikiSum (M) & 18.2 & 51.9 & 69.8 & 78.2 \\
    \hline
    SumREN (M) & 16.8 & 63.1 & 86.4 & 93.4 \\
    \hline
    \end{tabular}
    \caption{Percentage of novel $n$-grams in the reference summaries of different summarization datasets. (S) and (M) denote single- and multi-document summarization respectively. Numbers for SumREN are computed by averaging over the two reference summaries. 
    }
    \label{tab:dataset_novel_ngrams}
\end{table}


\subsection{Silver Training Data Generation}
\label{sec:silver-standard}
Given the cost associated with annotating statements and writing summaries, we automatically generate large-scale silver-standard training data for our task. 
Specifically, we leverage GPT-3 \cite{brown2020language} to automatically generate abstractive silver-standard summaries of the reported statements. This can be achieved by prompting \cite{liu2021pre}, which involves decomposing the task into an instruction (or a  `prompt') that is then provided to the model along with the input as the context. In our scenario, the input would be the reported statements and a speaker--automatically identified through the reported speech system that we build and describe in Section \ref{sec:reported_extraction} and the prompt would be \textit{``Summarize what $<$speaker$>$ said:''}. Similar to the gold-standard dataset, statements corresponding to the same speaker are grouped together before prompting GPT-3 to generate the summary. Overall, we generate 10,457 examples for our silver training set. 





\section{Models}

\begin{table*}[!htb]
\small
    \centering
    \def\arraystretch{1.2}
    \begin{tabular}{p{4.7em}|c|c|cccc|cccc}
    \hline
    \multicolumn{1}{c|}{\multirow{2}{*}{\textbf{Setting}}}&\multicolumn{1}{c|}{\multirow{2}{*}{\textbf{Model}}} & \multicolumn{1}{c|}{\multirow{2}{*}{\textbf{Approach}}} & \multicolumn{4}{c|}{\textbf{Dev}} &  \multicolumn{4}{c}{\textbf{Test}} \\
    & & & \textbf{R-1} & \textbf{R-2} & \textbf{R-L} & \textbf{BertScore} & \textbf{R-1} & \textbf{R-2} & \textbf{R-L} & \textbf{BertScore}  \\
    \hline
     \multirow{3}{*}{\parbox{4.7em}{\centering Baselines (Zero-shot)}} & GR-SUM & \multicolumn{1}{c|}{\multirow{4}{*}{QFS}} & 38.73 & 15.32 & 24.70 & 16.45 & 35.99 & 12.05 & 22.18 & 14.96\\
     & RelReg & & 35.40 & 11.88 & 22.97 & 21.64 & 31.49 & 8.38 & 20.02 & 17.24\\
     & SegEnc & & 38.53 & 14.99 & 24.98 & 26.26 & 36.62 & 11.77 & 22.99 & 23.26\\
     & \multicolumn{1}{c|}{GPT-3} &  & 42.34 & 16.71 & 29.12 & 34.08 & 39.45 & 13.78 & 26.72 & 31.16\\
    \hline
   \multirow{3}{*}{Zero-shot} & BART & Pipeline & 40.85 & 16.99 & 27.63 & 30.38 & 37.28 & 13.16 & 24.45 & 29.36\\
    
      
   & GPT-3 & Pipeline & 44.49 & 18.51 & 31.21 & \textbf{40.12} & 42.29 & 16.02 & 29.33 & \textbf{37.68}\\
     & \demph{GPT-3} & \demph{Pipeline (Oracle)} & \demph{47.27} & \demph{20.74} & \demph{33.98} & \demph{42.65} & \demph{45.45} & \demph{17.89} & \demph{31.27} & \demph{40.29}\\
    \hline
    
    \multirow{2}{*}{\parbox{4.7em}{\centering + Silver Training}} & SegEnc & QFS & 47.09 & 20.05 & 31.99 & 38.64 & 44.35 & 17.47 & \textbf{29.69} & 36.26\\
    & BART & Pipeline & 46.14 & 18.92 & 31.37 & 34.17 & 43.00 & 15.95 & 28.66 & 34.55\\
    \hline
    
   \multirow{3}{*}{\parbox{4.7em}{\centering + Gold Finetuning}} & SegEnc & QFS & \textbf{48.30} & \textbf{22.45} & \textbf{32.98} & 39.95 & \textbf{45.06} & \textbf{18.45} & 29.43 & 36.71\\
    & BART & Pipeline & 46.59 & 20.38 & 32.31 & 37.78 & 44.38 & 17.53 & 29.62 & 35.72\\
    & \demph{BART} & \demph{Pipeline (Oracle)}  & \demph{51.11} & \demph{24.23} & \demph{35.92} & \demph{42.28} & \demph{47.82} & \demph{20.23} & \demph{32.20} & \demph{39.61}\\
    \hline
    
  
    \hline
    \end{tabular}
    \caption{ROUGE and BertScore performance of various models on the SumREN benchmark. We explore both query-focused (QFS) and pipeline-based approaches under zero-shot, silver-training and gold-fine-tuning settings. \textit{Pipeline (Oracle)} corresponds to using the gold reported statements as input to the summarization model and is reported for the best setup for each of the zero-shot and fine-tuned models.}
    \label{tab:model_results}
\vspace{-0.75em}
\end{table*}
Here, we describe our proposed pipeline-based approach along with several strong baselines that we experiment with. 



\label{sec:query-focused}
\subsection{Query-Focused Summarization Baselines}
Our proposed task requires generating a summary of the reported statements, given a set of news articles and the name of the speaker as input. To leverage existing models, our reported-speech summarization task can be approached as query-focused summarization -- by  generating a summary of the given text conditioned upon a query.  Specifically, given the name of the speaker, the corresponding query can be formulated as: \textit{``Summarize what $<$speaker$>$ said.''}. Following this, we explore multiple query-focused summarization approaches, which we describe below.

\begin{itemize}[]
    \item \textbf{GR-SUM}~\cite{wan2008using} uses an unsupervised graph-based extractive method where each source sentence is treated as a node.\footnote{We use the source code from \citet{DBLP:journals/tacl/CMDPControlSum21}.} It uses a random-walk algorithm to rank the input sentences based on the adjacency weights and the topic relevance vectors for each node. 
    \item \textbf{RelReg}~\cite{vig2021exploring} uses a two-step process. First a regression model is used to extract a contiguous span within the input that is relevant to the input query. The extracted context is then passed along with the query to a BART model to generate a summary. Both the regression and BART models are trained on QMSum~\cite{DBLP:conf/naacl/QMSum21}, a query-focused meeting summarization dataset. 
    \item \textbf{SegEnc}~\cite{vig2021exploring} is an end-to-end generative model that first splits the source documents into overlapping text segments. Each of these segments is then concatenated with the input query and independently encoded by a Transformer encoder. The encoded segments are then concatenated into a sequence of vectors and fed into a Transformer decoder to generate the summary. The model is pre-trained on WikiSum dataset \cite{DBLP:conf/iclr/wikisum18} and finetuned on QMSum dataset \cite{zhong2021qmsum}. 
    \item \textbf{GPT-3}: In addition to these baselines, we also explore directly providing the source news articles as input to GPT-3  and using the query as the prompt.
\end{itemize}

\subsection{Pipeline-based Summarization Framework}
\label{sec:pipeline-based}
We utilitize a pipeline-based approach for summarizing reported speech. 
The proposed pipeline involves three main steps; (1) extracting reported statements and their speakers from the given set of news articles, (2) grouping statements together that come from the same speaker, and (3) generating a summary for each group of reported statements.

\subsubsection{Reported Speech Extraction:}
\label{sec:reported_extraction}
Given a collection of news articles and a speaker, we aim to identify all reported statements along with the corresponding speakers. To this end, we build a span-tagging system that leverages a Transformer-based encoder to identify the spans of statements and the corresponding speaker. The model is trained using the PolNeAR corpus \cite{newell2018attribution} which 
provides annotated triples of \textit{source} (i.e. speaker), \textit{cue} (i.e. words that indicate the presence of attribution), and \textit{content} (i.e. the statements made by the speaker) for statements made in the news. 

Given an input paragraph of length $T$, we use a BERT encoder to learn the representation $H \in R^{TXD}$ -- of hidden dimension $D$ -- for the input sequence. We then add a binary classification head to identify whether or not the input paragraph contains a reported statement and a $BIO$ sequence labeling head to identify the spans of the statement and the speaker. 
The binary classification $y^{cls}$ and the token label $Y^{span}_i\in R^K$ probabilities are calculated as follows:
\begin{align}
   y^{cls}  &= \sigma(w^{cls} \cdot H_{CLS} + b^{cls}) \\
   Y^{span}_i  &= \text{softmax}(W^{sp} H_i + b^{sp})
\end{align}
where $w^{cls}\in R^D$ and $W^{sp}\in R^{K\times D}$ are the weights, $b^{cls}$ and $b^{sp}$ are the bias terms, $K$ is the total number of $BIO$ tags, $H_{CLS}$ and $H_i$ denote the representation of the $CLS$ token and the $i$-th token respectively. 

Finally, the model is trained with a multi-task learning objective by using a joint loss that performs a weighted sum of the classification -- binary cross entropy (BCE) -- and the sequence labeling head -- Cross Entropy (CE) -- losses.
\begin{align}
    L = \alpha \cdot \textbf{} BCE(y^{cls}, \hat{y}^{cls}) + \beta \cdot  CE(Y^{sp}, \hat{Y}^{sp})
\end{align}
where $y^{cls}$ and $\hat{y}^{cls}$ correspond to the predicted and ground-truth classification label respectively, $Y^{sp}$ and $\hat{Y}^{sp}$ denote the predcited and ground-truth token labels respectively, $\alpha$ and $\beta$ are tunable hyper-parameters.\footnote{In our experiments, we set $\alpha$ to $1$ and $\beta$  to $0.4$.} 
\subsubsection{Speaker Co-reference Resolution:}
In order to group the statements by the speaker, we need to perform co-reference resolution since speakers can be referred to by different nominal (e.g., Biden, Joe Biden, Joe R. Biden) and pronominal (e.g., He) mentions. To achieve this, we utilize an existing information extraction system \cite{li2020gaia}, and updated it with a co-reference resolution from \citet{lai2022end}. As we show later, using co-reference resolution considerably increases the coverage of reported statements by a given speaker.
\subsubsection{Summary Generation:}
\label{sec:summarizing}
Given a set of reported statements for a speaker, we aim to generate a concise summary of the statements.
The summary generation process for the extracted reported statements of a given speaker is akin to single-document summarization. The reported statements are concatenated before getting passed as input to a BART \cite{lewis2020bart} model. The summarization model, trained on CNN-DailyMail \cite{hermann2015teaching}, is first used in a zero-shot setting. This model then undergoes silver-training and gold-finetuning, the details of which are provided in Section \ref{sec:train_setup}.



\section{Experiments}

\label{sec:experiments}

\subsection{Training Setup and Metrics}
\label{sec:train_setup}
We explore two methods for fine-tuning our base summary generation models: Silver Training and Gold Fine-tuning. 
During silver training, the models are fine-tuned on the silver-standard training data. For gold fine-tuning, we add a second fine-tuning step using the gold data. 

For evaluation, we use ROUGE \cite{lin2004rouge} 
and choose the best models based on ROUGE-L performance on the development set.\footnote{We use the \textsc{score\_multi} function from the \textsc{rouge\_score} python package: \url{https://pypi.org/project/rouge-score/}}
We also report BertScore \cite{bert-score} which leverages pre-trained contextual embeddings from BERT and matches words in candidate and reference sentences by cosine similarity. 
As opposed to ROUGE which measures the lexical similarity between the source and generated summaries, BertScore is capable of capturing the semantic similarity. 
\subsection{Results} 
Table \ref{tab:model_results} compares the performance of our proposed pipeline-based approach against the QFS baselines with and without fine-tuning using our silver and gold data.
For the baselines, GPT-3 performs best, justifying the choice of using it for generating silver-standard training data. We find that using silver training data for fine-tuning improves the performance of both the query-focused SegEnc and pipeline-based BART considerably, even outperforming GPT-3 in terms of ROUGE. Finally, we see that the models further benefit by fine-tuning on the gold human-annotated training data. 

We also find that using the pipeline approach, where we first extract the reported statements before passing them to GPT-3, achieves considerably better scores than passing the raw articles to GPT-3. However, GPT-3 has relatively lower ROUGE scores than smaller models (SegEnc and BART) that have been fine-tuned using gold data. We hypothesize that this could be attributed to the fact that GPT-3 generates more abstractive summaries (as will be shown in Table \ref{tab:generated_ngrams}) thereby leading to higher scores for metrics that are capable of capturing semantic similarity.

In zero-shot settings, the pipeline-based model considerably outperforms query-focused baselines, showing the benefit from explicitly extracting reported statements. However, in both silver training and gold fine-tuning settings, the SegEnc model consistently outperforms the pipeline-based models, suggesting that it may be possible to implicitly identify reported statements within an end-to-end approach. Nevertheless, when using the oracle reported statements, BART surpasses SegEnc -- implying that employing better reported speech and co-reference resolution systems will considerably improve the pipeline-based approach.

\subsubsection{Reported Speech Extraction Performance\\}
Next, we analyze the performance of the proposed reported speech extraction component to identify areas of improvement. 
We compare our span tagging approach with a Semantic Role Labeling (SRL) baseline to identify reported statements and the corresponding speakers and evaluate using character-level offset \fscore{} of the extracted span. SRL outputs the verb predicate-argument structure of a sentence such as who did what to whom. Given a paragraph as an input, we filter out verb predicates matching a pre-defined set of cues that signal attribution (e.g., \emph{say}, \emph{believe}, \emph{deny}) and identify these sentences as containing reported statements.\footnote{Full list of used cues is provided in the appendix.} The sentences encompassing ARG-1 of the predicate are considered as the reported statement and the span corresponding to ARG-0 (agent) is used as the speaker.

\begin{table}[!htb]
\footnotesize
    \centering
    \def\arraystretch{1.2}
    \begin{tabular}{c|ccc|ccc}
    \hline
    \multicolumn{1}{c|}{\multirow{2}{*}{\textbf{Model}}} & \multicolumn{3}{c|}{\textbf{Dev}} & \multicolumn{3}{c}{\textbf{Test}} \\
    & \textbf{P} & \textbf{R} & \textbf{F1} & \textbf{P} & \textbf{R} & \textbf{F1} \\
    \hline
    SRL& 84.3 & 42.7 & 56.7 & 83.3 & 40.8 & 54.8\\
    + co-reference & 82.2 & 68.1 & 74.5 & 83.1 & 68.3 & 75.0\\
    \hline
    Span Tagging & 80.3 & 48.6 & 60.5 & 80.2 & 45.0 & 57.6\\
    + co-reference & 78.7 & 69.9 & 74.1 & 78.2 & 73.0 & 75.5\\

    \hline 
    \end{tabular}
    \caption{Performance (in \%) of different approaches for identifying reported statements corresponding to a given speaker for the summaries in SumREN. ``\textit{+ co-reference}'' corresponds to adding co-reference resolution for the speaker mention extracted by the system.}
    \label{tab:summary_statements}
\end{table}

As shown in Table \ref{tab:summary_statements}, our proposed span tagging model outperforms SRL, especially in terms of recall which ensures better coverage of information for the summarization step. We also find that incorporating co-reference resolution for speaker identification considerably improves recall with almost the same or slightly lower precision.
Table \ref{tab:all_statements} measures the performance of the proposed span-tagging approach for speaker extraction against the SRL baseline. We report both string exact-match and \fscore{}, both of which are commonly used in extractive question answering \cite{rajpurkar2016squad}.
We find that the performance of different approaches for identifying the speaker of a given reported statement improves significantly when using co-reference resolution. This is crucial for correctly grouping statements from the same speaker together. 

\begin{table}[!htb]
\footnotesize
    \def\arraystretch{1.2}

    \centering
    \begin{tabular}{c|cc|cc}
    \hline
    \multicolumn{1}{c|}{\multirow{2}{*}{\textbf{Model}}} & \multicolumn{2}{c|}{\textbf{Dev}} & \multicolumn{2}{c}{\textbf{Test}} \\
    & \textbf{Exact-Match} & \textbf{F1} & \textbf{Exact-Match} & \textbf{F1} \\
    \hline 
    SRL & 20.8 & 48.1 & 16.1 & 44.4\\
    + co-reference & 62.9 & 73.7 & 69.1 & 77.1 \\
    \hline
    Span Tagging & 22.3 & 51.2 & 18.8 & 49.3 \\
    + co-reference & 63.3 & 74.8 & 69.8 & 78.4 \\
    \hline
    
    \end{tabular}
    \caption{Performance (in \%) of the proposed span tagging component -- against the baseline -- on identifying the speakers corresponding to the given reported statements with and without co-reference resolution. 
    }
    \label{tab:all_statements}
\vspace{-0.75em}
\end{table}
%
\subsubsection{Parameter-Efficient versus Direct Fine-tuning\\} 
In addition to full fine-tuning methods, we also explore leveraging parameter-efficient fine-tuning approaches to directly fine-tune on the small-scale gold training data. We use LORA \cite{hu2021lora}, an efficient fine-tuning technique that injects trainable low-rank decomposition matrices into the layers of a pre-trained model. Table \ref{tab:pe_approaches} compares the performance of three different fine-tuning strategies, namely \emph{Full FT} (silver training + gold fine-tuning), \emph{Gold FT} (direct gold fine-tuning) and PE FT (parameter-efficient gold fine-tuning). 
We find that the benefit of incorporating the silver-standard training data can be seen from the fact that Full FT considerably outperforms Gold FT. We also observe that \emph{PE FT} with LORA, which fine-tunes only 0.3\% of model parameters, can achieve a comparable performance to Full FT while also consistently outperforming \emph{Gold FT}. 
This shows that parameter-efficient fine-tuning is effective for our pipeline-based reported speech summarization framework, with future work potentially benefiting from better PE approaches \cite{liu2021p}. 
\begin{table}[!htb]
\footnotesize
    \def\arraystretch{1.2}
    \centering
    \begin{tabular}{l|cc|cc}
    \hline
    \multicolumn{1}{c|}{} &  \multicolumn{2}{c|}{\textbf{Dev}} &  \multicolumn{2}{c}{\textbf{Test}} \\
    & \textbf{R-1/2/L} & \textbf{BertS} & \textbf{R-1/2/L} & \textbf{BertS}  \\
    \hline
    Full FT & 51.1/24.2/35.9 & 42.3 & 47.8/20.2/32.3 & 39.6\\
    \hline
    Gold FT & 50.0/23.9/35.1 & 40.9 & 47.2/19.5/31.4 & 38.4\\
    \hline
    PE FT & 50.7/24.6/36.1 & 42.0 & 47.8/20.2/31.8 & 39.4\\
    \hline
    \end{tabular}
    \caption{Comparison of performance of parameter-efficient fine-tuning for BART when used for summarization with oracle reported statements. \textit{Full FT} corresponds to silver training + gold FT.}
    \label{tab:pe_approaches}
\end{table}

\begin{table*}[]
\scriptsize
\def\arraystretch{1.4}
    \centering
    \begin{tabular}{p{69em}}
    \multicolumn{1}{c}{\textbf{Event}: 2017 Solar Eclipse \hspace{1.5em} \textbf{Speaker}: Jamie Yuccas \hspace{1.5em} \textbf{News Article}: \url{https://www.cbsnews.com/news/solar-eclipse-august-21-2017-across-america}}\\
    \hline
    \hspace{30em}\textbf{Gold Reported Statements}\\
    \hline
    ``They're expecting about a million people to enter the state, a million out-of-towners are supposed to come to the state of Oregon,'' said CBS News correspondent Jamie Yuccas. ``Where we're located in Madras, they're expecting between 100,000 and 200,000 people.''\\
    \hdashline
    She said the local residents have been ``really, really nice and accommodating.''\\
    \hdashline
    ``What the mayor said to me was kind of funny,'' Yuccas said. ``He said 'you know, I think it's going to be one of those situations that you might not get your newspaper, you might not have your daily Starbucks and if that happens, I guess it's a first-world problem, and you're going to have to figure out your own survival skills.'''\\
    \hdashline
    She laughed, saying ``there are going to be some minor inconveniences, but I actually think they had a pretty good plan together.''\\
    \hline
    \textbf{SegEnc (QFS):} Jamie Yuccas has said that the local residents of Madras, Oregon have been very accommodating towards people who are coming to see the solar eclipse. She laughed when she heard that the mayor of the town thought it would be a first-world problem if the eclipse did not occur. She said that they are expecting about a million people to enter the state, and that a million out-of-towners are supposed to come to the state. \\
    \hline
    \textbf{GPT-3 (QFS):} Jamie Yuccas is a CBS News correspondent who is reporting from Madras, on the upcoming solar eclipse. She says that the city is expecting between 100,000 and 200,000 visitors for the event, and that the locals have been very accommodating. Yuccas also says that the eclipse is expected to be the most observed eclipse in history.\\
    \hline
    \textbf{GPT-3 (Pipeline):} Jamie Yuccas said that the city of Portland is expecting about a million out-of-towners to come to Oregon for the eclipse, and that the locals have been very accommodating. She said that the mayor told her that there may be some minor inconveniences, but that they have a good plan in place.\\
    \hline
    \textbf{BART (Pipeline):} According to the correspondent, the state of Oregon is expecting a million people from out-of-towns to come to the state, and the local residents have been very nice and accommodating. However, there will be some minor inconveniences, but the state had a good plan in place.\\
    \hline
    \textbf{Gold:} Jamie Yuccas said that a million tourists are supposed to come to the state of Oregon and between 100,000 and 200,000 people are expected in Madras, where she is located. She also mentioned minor inconveniences could occur derived from the event, but overall, they had a good plan set in place.\\
    \hline
    \end{tabular}
    \caption{Model outputs for an example in SumREN, along with the gold reported statements. Summaries from the QFS approaches contain factually inconsistent fragments, while those from  pipeline-based
approaches better match the gold summary.}
    \label{tab:analysis}
    \vspace{-0.95em}
\end{table*}

\subsubsection{Abstractiveness and Factuality of Generated Summaries \\}
We investigate the effect of using silver and gold data for fine-tuning, on both the abstractiveness and factuality of generated summaries. There is generally a trade-off  between abstractiveness and factual consistency of the summary against the source input~\cite{dreyer2021analyzing}. Hence, the goal of any abstractive summarization system is to generate more abstractive summaries while maintaining a high level of factual consistency with the source.
\begin{table}[!htb]
\footnotesize
\def\arraystretch{1.2}
    \centering
    \begin{tabular}{c|c|c|cccc}
    \hline
     & \textbf{Model} & \textbf{Setting} & \textbf{Uni} & \textbf{Bi} & \textbf{Tri} & \textbf{MINT}\\
    \hline
    \parbox[t]{2mm}{\multirow{4}{*}{\rotatebox[origin=c]{90}{QFS}}} & \multicolumn{1}{c|}{\multirow{3}{*}{SegEnc}} & Zero-Shot & 1.0 & 6.6 & 13.1 & 11.1\\
   & & + Silver Train & 2.8 & 22.8 & 39.3 & 31.2\\
    & & + Gold FT & 3.6 & 26.6 & 46.6 & 38.4\\
    \cline{2-7}
    & GPT-3 & Zero-Shot & 3.8 & 26.2 & 44.2 & 38.9\\
    \hline
    \parbox[t]{2mm}{\multirow{4}{*}{\rotatebox[origin=c]{90}{Pipeline}}} & \multicolumn{1}{c|}{\multirow{3}{*}{BART}} & Zero-Shot & 1.9 & 11.5 & 20.5 & 15.3\\
   & & + Silver Train & 3.3 & 24.8 & 41.6 & 32.9\\
    & & + Gold FT & 4.7 & 30.6 & 52.1 & 43.5\\
    \cline{2-7}
    & GPT-3 & Zero-Shot & 5.7 & 35.2 & 56.6 & 49.6\\
    \hline
    \end{tabular}
    \caption{Abstractiveness and novelty scores -- measured by \% of novel ngrams -- of the generated summaries using silver and gold data for fine-tuning the models. The novelty is computed with respect to the source news articles.}
    \label{tab:generated_ngrams}
    \vspace{-0.3em}
\end{table}

For abstractiveness, we measure it through the percentage of novel $n$-grams (uni, bi and tri-grams), as well as MINT (\emph{M}etric for lexical \emph{IN}dependence of generated \emph{T}ext) \cite{dreyer2021analyzing} which is computed based on the n-gram precision and longest common sub-sequence length of the generated summary. 
As shown in Table \ref{tab:generated_ngrams}, we find that models in zero-shot settings are considerably more extractive, and that abstractiveness of generated summary significantly increases from both silver training and gold fine-tuning. 
Further, we notice that our pipeline-based approach is considerably more abstractive than the QFS approach, demonstrating that incorporating an explicit statement extraction component helps the summarization model focus on paraphrasing and synthesizing the selected statements into the summary.

\begin{table}[!htb]
\footnotesize
\def\arraystretch{1.2}
    \centering
    \begin{tabular}{c|c|ccc}
    \hline
    \textbf{Approach} & \textbf{Model} & \textbf{FactCC} & \textbf{Entity P} & \textbf{MINT}\\
    \hline
    \multicolumn{1}{c|}{\multirow{2}{*}{QFS}} & GPT-3 & 45.4  & 61.7 & 38.9 \\
    
   & SegEnc  & 50.8  &  \textbf{75.4} & 38.4\\
    \hline
    \multicolumn{1}{c|}{\multirow{2}{*}{Pipeline}} & GPT-3 & 50.2 & 73.2 & \textbf{49.6}\\

   & BART  & \textbf{52.1} & 74.6 & 43.5\\
    \hline
    \multicolumn{1}{c|}{\multirow{2}{*}{\demph{Pipeline (Oracle)}}} & \demph{GPT-3} & \demph{52.0}  & \demph{78.9} & \demph{51.3}  \\
 
   & \demph{BART} & \demph{55.0} & \demph{84.6} & \demph{44.0}  \\
    \hline
    \end{tabular}
    \caption{Comparison of factuality (measured by FactCC and Entity Precision) of generated summaries relative to abstractiveness (measured by MINT). Models considered are after silver train + gold FT, except GPT-3 which is not fine-tuned.}
    \label{tab:factuality}
    \vspace{-0.3em}
\end{table}

\begin{table*}[]
\scriptsize
\def\arraystretch{1.4}
    \centering
    \begin{tabular}{p{69em}}
    \multicolumn{1}{c}{\textbf{Event}: Disappearance of journalist Jamal Khashoggi \hspace{4.5em} \textbf{Speaker}:  Robert Mahoney} \\
    \hline
    \hspace{30em}\textbf{Gold Reported Statements}\\
    \hline
    \textbf{Statement 1:} ``The worrying trend that we see is a crackdown on critical and independent journalism in Saudi and about Saudi,'' Robert Mahoney, Deputy Executive Director of CPJ, told CBS News. ``It is a very repressive place.''\\
    \hdashline
    \textbf{Statement 2:} He said that there is no surprise that Riyad is looking to quash criticism, but this case is beyond the sad but true usual jailing of journalists.\\
    \hdashline
    \textbf{Statement 3:} ``This is an incredible escalation if these stories are true,'' Mahoney said. ``We have had cases where journalists are killed in another country, but no one as prominent as this, not a U.S. resident who was writing for the Washington Post. He is a journalist but he was also from the Saudi establishment before the present Crown Prince came.''\\
    \hdashline
    \textbf{Statement 4:} Mahoney said that the disappearance of Khashoggi is ``designed in a very brutal way to send a message `if you speak out against the Saudi government, if it could happen to him, it could happen to you.'''\\
    \hdashline
    \textbf{Statement 5:} ``With no body, there is no crime ... is the way the Saudis are treating the case right now'', Mahoney said. \\
    \hdashline
    \textbf{Statement 6:} The Committee to Protect Journalists Deputy Executive Director Robert Mahoney said in a statement that the group was ``alarmed'' by the reports that Khashoggi may have been killed inside the consulate.\\
    \hdashline
    \textbf{Statement 7:} ``The Saudi authorities must immediately give a full and credible accounting of what happened to Khashoggi inside its diplomatic mission,'' Mahoney said. ``The country has stepped up its repression of critical journalists in the past year at home. We hope this has not now spread abroad.''\\
    \hline
    \textbf{BART (Pipeline):} Robert Mahoney, Deputy Executive Director of CPJ, said that the disappearance of journalist Jamal Khashoggi was a result of a crackdown on critical and independent journalism in Saudi Arabia, and that if the reports of his disappearance are true, it would be an escalation beyond the usual jailing of journalists. Mahoney expressed concern that the Saudi authorities must give a full and credible account of what happened to the journalist inside the consulate. \\
    \hline
    \textbf{Human:} Robert Mahoney said the reports of Jamal Khashoggi's disappearance alarmed the group, standing as the most prominent case of the sad but real jailing and killing of journalist in an attempt to crack down critical and independent journalism in and about Saudi designed brutally as a message to other journalists. Mahoney demands a full account from Saudi authorities of Khashoggi's whereabouts despite their evasive approach towards the case claiming the absence of a body.\\
    \hline
    \end{tabular}
    \caption{Example output from the proposed BART-based pipeline, along with the gold reported statements and corresponding reference summary. The human summary has considerably higher coverage of the input statements than the BART summary.}
    \label{tab:coverage_example_1}
\end{table*}

For factuality, we use FactCC \cite{kryscinski2020evaluating}, which \citet{pagnoni2021understanding} show to correlate most with human factuality labels. In addition, Entity Precision~\cite{DBLP:conf/eacl/NanNWSZZMX21} is calculated based on the percentage of named entities in the generated summary that are present in the gold reported statements.
In Table \ref{tab:factuality}, we observe that while our proposed pipeline-based approach is considerably more abstractive than the QFS baselines, it still maintains high entity precision and a slightly higher FactCC score. As expected, we see that using gold (oracle) statements as input to the summarization step improves the factual consistency scores.

\subsubsection{Human Evaluation\\}

We also performed a human study of the summaries generated using GPT-3 via both pipeline-based and QFS approaches. We chose GPT-3 summaries since they have consistently high scores across Rouge-L, BertScore and abstractiveness. Annotators were presented with the summaries along with the ground-truth reported statements, and were asked to evaluate on a scale of 1-3 for factual consistency, informativeness and coherence\footnote{Detailed guidelines are provided in the appendix.}. Evaluation for \textit{factual consistency} involves looking for major or minor factual errors in the summary, \textit{informativeness} is about how well the summary expresses the main point of the reported statements, and \textit{coherence} is mainly checking whether the summary has a good flow and facts are presented in a logical order. The annotations were crowd-sourced via MTurk. Table \ref{tab:human_study} shows results from the human study. We find that summaries from the pipeline-based approach have considerably better factual consistency with the ground-truth reported statements, with slight improvements in informativeness. Concurring with recent observations \cite{goyal2022news, zhang2023benchmarking} on the quality of summaries from large language models, we see that the summaries based on the two approaches, which both come from GPT-3, are very coherent.

\begin{table}[!htb]
\footnotesize
\def\arraystretch{1.2}
    \centering
    \begin{tabular}{c|ccc}
    \hline
    \textbf{Approach} & \textbf{Consistent} & \textbf{Informative} & \textbf{Coherent}\\
    \hline    
    GPT-3 (QFS) & 2.76 & 2.92 & 2.99\\
    GPT-3 (Pipeline) & 2.92 & 2.95 & 2.99\\    
    \hline
    \end{tabular}
    \caption{Results from human study on summaries from GPT-3 via pipeline-based and QFS approaches, when evaluated for factual consistency, informativeness and coherence.}
    \label{tab:human_study}
    \vspace{-1.6em}
\end{table}

\subsection{Manual Error Analysis}

Table \ref{tab:analysis} shows outputs from different models for an example in \textsc{SumREN}. We see that summaries from the query-focused approaches contain factually inconsistent fragments: SegEnc output suggests that the mayor \textit{``thought it would be a first-world problem if the eclipse did not occur''} whereas the mayor actually refers to \textit{``people not getting their newspapers or their daily Starbucks''} as the first-world problems; GPT-3 (QFS) misattributes the statement \textit{``the eclipse is expected to be the most observed eclipse in history''}. On the other hand, summaries from pipeline-based approaches match the gold summary better, with those from BART and GPT-3 (Pipeline) being fairly similar in quality.



We also analyzed some of the errors made by the reported speech extraction component of the proposed pipeline. As Table \ref{tab:model_results} shows, there is still a considerable room for improving our pipeline-based approach with better reported speech extraction systems. We found that the same entity can be referred to by different aliases that the co-reference system sometimes fails to capture (e.g., ``Islamic State'' and ``ISIS'' or ``Anthony M. Kennedy'' and ``Justice Kennedy''). 
Utilizing entity-linking \cite{ayoola2022refined} will likely improve co-reference for entities with different aliases. In addition, we found that spelling variations; e.g., Nikos vs. Nicos, Muhammad vs. Mohammed or Sergey vs. Sergei, were also frequently missed by the system. We believe that incorporating character-level features into the co-reference resolution system will make it more robust to such variations.


Finally, to analyze the informativeness of the generated summaries, we calculated the percentage of input reported statements covered in the output summary. To obtain alignments between source-summary pairs, we leveraged SuperPAL \cite{ernst2021summary} which aligns OpenIE-extracted \cite{stanovsky2018supervised} propositions in the generated summary with those in the source sentences. We found 
that human summaries cover considerably more percentage of the input reported statements (57.5\%) compared to summaries from BART (51.2\%) and GPT-3 (46.5\%) in \textit{Pipeline (Oracle)} settings. Table \ref{tab:coverage_example_1} shows one such qualitative example where the human summary covers information from statements 1, 2, 4, 5, 6 and 7, whereas the BART model output only covers statements 1, 2, 3 and 7. 
In order to explicitly improve coverage, future work can explore incorporating more control into the generation output by clustering the salient spans within the reported statements and separately generating summaries for each cluster, similar to \citet{ernst2022proposition}.

\section{Conclusion \& Future Work}
In this work, we introduce a new challenging task of summarizing reported speech in news and release \textsc{SumREN} to promote more research in this direction. 
We propose a pipeline-based framework for summarizing reported statements and show that the proposed approach can generate summaries that are both more factual and abstractive than QFS. 
Future work involves improving reported speech extraction performance by leveraging entity-linking and by incorporating character-level features for speaker co-reference resolution. Another direction is to improve the coverage of salient spans in reported statements by adding more explicit control into the generation process. 


\section{Acknowledgment}

We would like to thank members of the Alexa Web Information team, especially Markus Dreyer and Sandeep Atluri, for useful discussions and feedback. We would also like to thank the anonymous reviewers for their insightful comments.

\bibliography{aaai23}
\clearpage
\appendix

\section{Appendix}
\label{sec:appendix}

\subsection{Annotation Process}

In this section, we provide more details about our annotation process for constructing the \textsc{SumREN} benchmark. Firstly, we describe our annotation interface for identifying reported statements and their corresponding speakers. Next, we expand on the definition of salient spans within reported statements and provide examples for how they are combined into a summary.

\subsubsection{Identifying Reported Statements} 
\begin{figure*}
    \centering
    \includegraphics[width=0.9\linewidth]{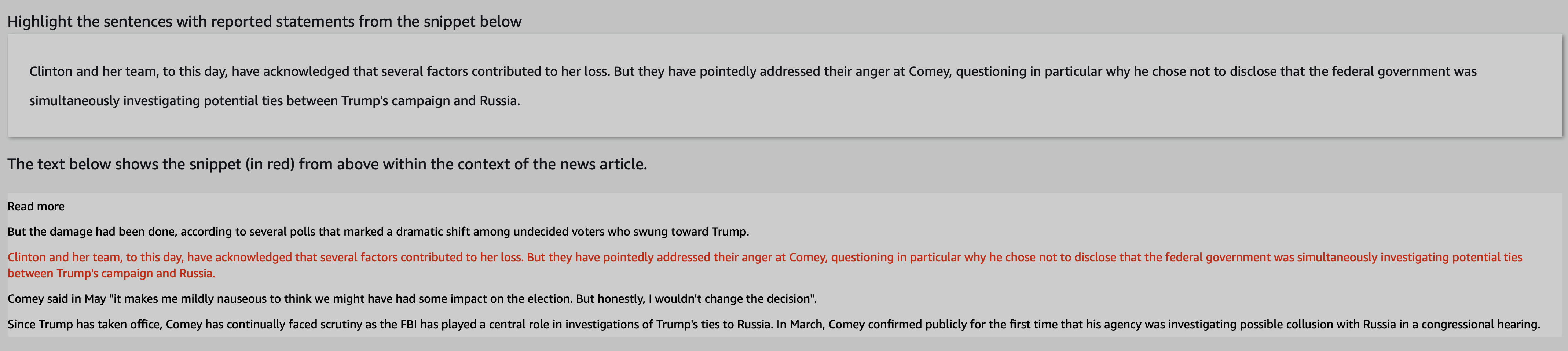}
    \caption{Annotation interface for the task of identifying sentences containing reported statements.}
    \label{fig:reported_statements_interface}
\end{figure*}

Figure \ref{fig:reported_statements_interface} shows a screenshot of the annotation interface for the task of identifying reported statements within news articles. Given a news snippet, annotators are asked to identify sentences within the snippet that contain reported statements. Reported statements involve both direct and indirect quotations of statements made by people or organizations. Direct quotations are statements that are within quotes. Indirect quotations usually do not use quotation marks and state what a person or organization has said.  Contiguous sentences that contain statements from the same speaker are considered to be corresponding to the same reported statement. The guidelines provided to the annotators are given below, with Figure \ref{fig:reported_statements_interface_ex} showing some sample annotations that were also shown in the interface.

\begin{itemize}
    \item Read the snippet carefully.

    \item Highlight the sentence that contains a reported statement.

    \item Multiple contiguous sentences from the same speaker correspond to a single reported statement.

    \item If the snippet contains statements from different speakers, annotate them separately with different labels. Use the labels incrementally i.e Statement 1 then 2 then 3.

    \item Please refer to the additional context around the snippet (provided in the white box below the snippet) if it's unclear whether or not the snippet contains a reported statement.
\end{itemize}

\begin{figure}[!htb]
    \centering
    \includegraphics[width=0.8\linewidth]{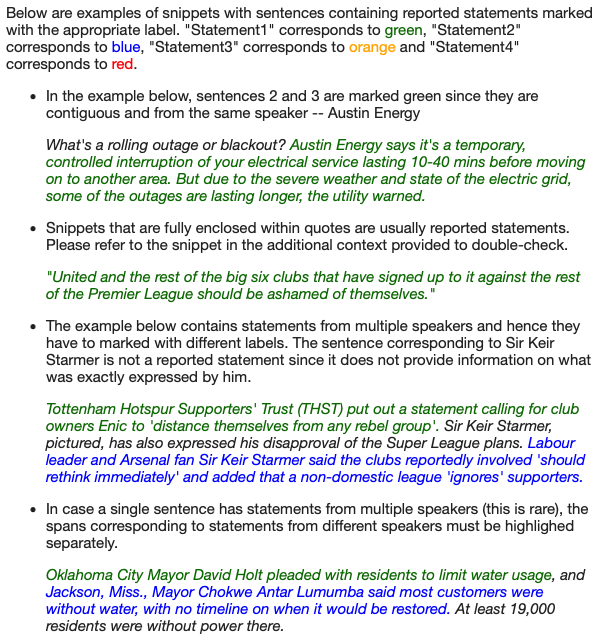}
    \caption{Sample annotations for reported statements that were also shown in the annotation interface.}
    \label{fig:reported_statements_interface_ex}
\end{figure}

\subsubsection{Annotating Speakers for Reported Statements} 
\begin{figure*}
    \centering
    \includegraphics[width=0.8\linewidth]{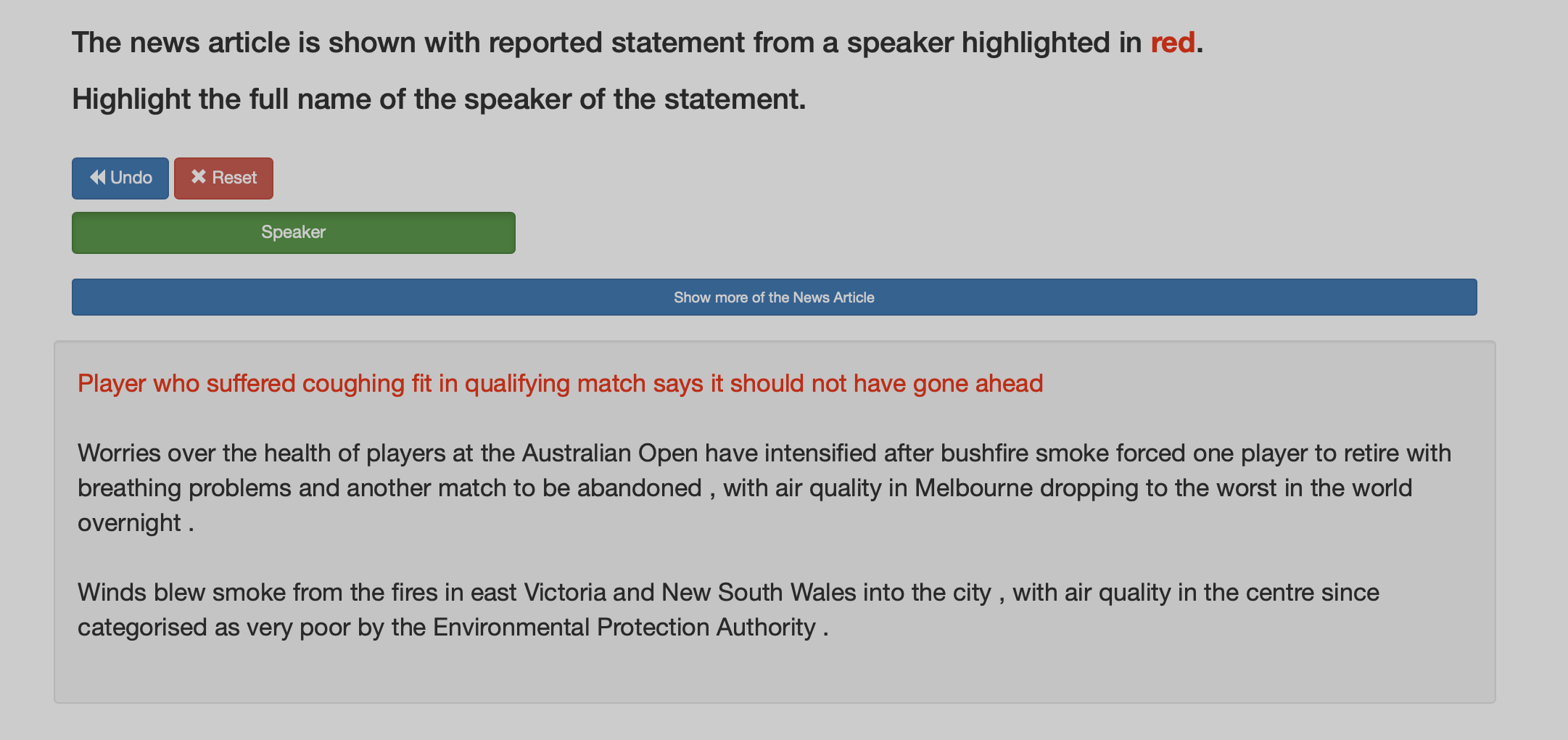}
    \caption{Annotation interface for the task of identifying the speaker for a given reported statement (highlighted  in red).}
    \label{fig:speaker_interface}
\end{figure*}

Given a reported statement, it is necessary to also identify the speaker of the statement. The speaker for a reported statement in news can be a person (e.g. Anthony Fauci, Donald Trump, etc.), an unnamed group of people (e.g. law makers, official), or an organization (i.e., a named group, e.g. such as World Health Organization, White House). 
Figure \ref{fig:speaker_interface} shows the screenshot of the speaker annotation interface where annotators are asked to identify the full name of the speaker for the reported statement that has been marked in red. This requires doing co-reference resolution and the full name needs to be marked within the news article that the statement belongs to. If the full name of the speaker is not mentioned (i.e. a pronoun, or only part of the name is present) within the shown snippet, the annotators would need to identify the full name the full name from beyond the snippet. Some instructions for identifying the full name include:

\begin{itemize}
    \item Salutations -- e.g.,  Mr., Mrs., Dr., etc. -- are not part of the full name. So for ``\emph{Dr. Anthony Fauci}'', the full name is  \emph{``Anthony Fauci"}.
    
    \item Titles and positions of people are not part of the full name. So if the text mentions ``\emph{Lawrence Gostin, a public health law expert from Georgetown University}, the full name would be ``\emph{``Lawrence Gostin}''.

    \item If the speaker is an organization and only the acronym is mentioned within the snippet, the annotator needs to look for whether the expansion of the acronym is present in the news article.
\end{itemize}

\begin{table}[t]
\scriptsize
    \centering
    \begin{tabular}{p{32em}}
    \hline
      Within hours of the U.S. Capitol being secured from a mob of pro-Trump supporters, \textcolor{red}{demonstrators took to the streets} in and \textcolor{red}{around Trump-named buildings}, such as those \textcolor{red}{in New York City, Chicago and Washington D.C.}, police said and photos show. \\
      \hline
      While \textcolor{red}{Republicans look inward} at the aftermath of the Capitol Hill riots after President Trump's address Wednesday, \textcolor{red}{Democrats are adding to the division}, Fox News contributor Charles Hurt told Fox \& Friends. \\
      \hline
      Musk said last month that the \textcolor{red}{Nasa money will help development of Starship}, which is meant to eventually launch atop a Super Heavy booster. \\
      \hline
    \end{tabular}
    \caption{Examples of reported statements along with their salient spans (highlighted in red).}
    \label{tab:salient_spans}
\end{table}

\begin{table*}[!htb]
\scriptsize
\def\arraystretch{1.4}
    \centering
    \begin{tabular}{p{69em}}
    \multicolumn{1}{c}{\textbf{Event}: Asian women shot in Atlanta \hspace{4.5em} \textbf{Speaker}: Scott Lee}\\
    \hline
    \multicolumn{1}{c}{\textbf{Gold Reported Statements}}\\
    \hline
    \textbf{Statement 1:} She spent most of her life in the New York City metropolitan area and moved to Atlanta several years ago to be closer to friends, said Scott Lee, her son-in-law.\\
    \hdashline
    \textbf{Statement 2:} ``She just liked to work,'' Lee said in an interview. ``It wasn't for the money. She just wanted a little bit of work for her life.''\\
    \hdashline
    \textbf{Statement 3:} ``Outside of the spa, Park was fit and active, Lee said.\\
    \hdashline
    \textbf{Statement 4:} ``She was very healthy,'' he said. ``Everybody said she was going to live past 100 years old.''\\
    \hdashline
    \textbf{Statement 5:} Lee said he had grown close to Park since marrying her daughter a decade ago. They lived under the same roof in Lyndhurst, N.J., before she relocated to Georgia. She was a dancer when she was younger, he said, and sometimes, the two of them would dance together. When they spoke, she made a point to refer to him by his given name rather than his in-law title in Korean - a reflection of their deep bond, he said.\\
    \hdashline
    \textbf{Statement 6:} She was planning to move back into Lee's home in June when her apartment lease in Atlanta ended, according to Lee.\\
    \hdashline
    \textbf{Statement 7:} Lee recalled: ``We always said to her, `Come back and be with us.' ''\\
    \hline
    \hline
    \textbf{BART (Pipeline):} Scott Lee, the son-in-law of Park Lee Park, said that Park was a hard worker who always wanted to work, even though she didn't have a lot of money, because she wanted a little bit of work for her life. Lee said that she was a dancer when she was younger and used to dance with her daughter, and they used to live together in Lyndhurst, N.J.\\
    \hline
    \textbf{Human:} Scott Lee remembered his mother-in-law, Park, who moved to Atlanta to get closer to her friends. Park was very active and very healthy, everyone believed that she would live long. Lee got close to Park when she married her daughter, and they created a deep bond, so she was always welcome when she came to visit him.\\
    \hline
    \end{tabular}
    \caption{An example where the human summary covers considerably more reported statements than the model output.}
    \label{tab:coverage_example_2}
    \vspace{-0.75em}
\end{table*}

\subsubsection{Salient Spans in Reported Statements} 
Compressing multiple reported statements into a summary requires being able to identify salient spans within these reported statements. A salient span is a sub-span within the reported statement that contains the crux of the information presented in the statement. Identifying salient spans can be seen as selecting the information that would be useful for composing the summary. Other parts of the statement usually expand on or provide context to the salient part. 

Table \ref{tab:salient_spans} shows some examples of reported statements along with their corresponding salient spans. In the first example, the initial part of the statement \textit{``Within hours of the U.S. Capitol being secured from a mob of pro-Trump supporters''} provides additional context, with the main intention of the statement being to convey that demonstrators took to the streets in these cities. In the third example, the part \textit{``which is meant to eventually launch atop a Super Heavy booster''} only expands on the main information around NASA providing funding. 

Tables \ref{tab:sum_ex_1} and \ref{tab:sum_ex_2} show examples for combining the salient spans within reported statements into a summary.

\begin{table}[!htb]
\renewcommand{\arraystretch}{0.9}
\scriptsize
    \centering
    \begin{tabular}{|p{32em}|}
                 \hline
                 \multicolumn{1}{|c|}{\textbf{Event: \textit{Capitol Hill Riots} \hspace{4em}Speaker: \textit{Police}}} \\
                 \hline
                 \hline
                 \multicolumn{1}{|c|}{\textbf{Reported Statements}}\\
                 \hline
                 \textcolor{red}{The riot left five dead}, including one pro-Trump demonstrator who was shot and killed by Capitol Hill police, and \textcolor{red}{at least 68 arrests}, according to D.C. police.  \\
                 \hline
                 Washington D.C.'s Metropolitan Police said Thursday morning \textcolor{red}{four people died and at least 68 people} were arrested in connection with the unrest of curfew violations.\\
                 \hline
                 Within hours of the U.S. Capitol being secured from a mob of pro-Trump supporters, \textcolor{red}{demonstrators took to the streets} in and around Trump-named buildings, such as those \textcolor{red}{in New York City, Chicago and Washington D.C.}, police said and photos show. \\
                 \hline
                 The New York Police Department was out in full force ahead of any potential unrest, another video shows. \textcolor{red}{Officers ultimately arrested nine people}, issuing summonses to seven before letting them go, police said.\\
                 \hline
                 Washington D.C.'s Metropolitan Police said \textcolor{red}{late Wednesday that by day's end, four people died and at least 52 people were arrested}. At least 14 MPD officers were hurt.  \\
                 \hline
                 \hline
                 \textbf{Summary:} \textit{D.C police said that by Thursday morning, four people had died and 68 were arrested, with at least 52 people arrested by late Wednesday. Demonstrators took to the streets around Trump-named buildings in New York, Chicago and Washington D.C, with nine people being arrested in New York.}\\
                 \hline
\end{tabular}
    \caption{An example showing the salient spans (highlighted in red) within reported statements and the summary.}
    \label{tab:sum_ex_1}
\end{table}

\begin{table}[!htb]
\renewcommand{\arraystretch}{0.9}
\scriptsize
    \centering
    \begin{tabular}{|p{32em}|}
                 \hline
                 \multicolumn{1}{|c|}{\textbf{Event: \textit{SpaceX's successful Starship mission} \hspace{2em}Speaker: \textit{Elon Musk}}} \\
                 \hline
                 \hline
                 \multicolumn{1}{|c|}{\textbf{Reported Statements}}\\
                 \hline
                 Musk tweeted \textcolor{red}{the landing was “nominal”} – by the book, in other words.\\
                 \hline
                 Musk said last month that \textcolor{red}{the Nasa money will help development of Starship}, which is meant to eventually launch atop a Super Heavy booster. He said it had been a ``pretty expensive'' project so far and mostly funded internally.\\
                 \hline
                 ``As you can tell, if you've been watching the videos, \textcolor{red}{we've blown up a few of them}. So excitement guaranteed, one way or another,'' Musk told reporters after the private company's second crew flight on 23 April.\\
                 \hline
                 \textcolor{red}{``Starship landing nominal''}, Musk tweeted after the landing.\\
                 \hline
                 \textcolor{red}{``Starship landing nominal''}, SpaceX Chief Executive Elon Musk said in a tweet shortly after the test.\\
                 \hline
                 Musk said last year that \textcolor{red}{the company could launch an unmanned mission to the Red Planet as soon as 2022}, with a possible crewed mission taking off two years after that.\\
                 \hline
                 Musk said in March that \textcolor{red}{SpaceX would use the Super Heavy booster to launch the massive Starship spacecraft into orbit} in a future test later this year.\\
                 \hline
                 \hline
                 \textbf{Summary:} \textit{Elon Musk tweeted that the Starship landing was nominal. He said that Nasa money will help develop the Starship, which would use a Super Heavy Booster to launch into orbit in future tests. In addition, SpaceX could launch unmanned and crewed missions to the Red Planet in the next couple of years.}\\
                 \hline
\end{tabular}
    \caption{An example showing the salient spans (highlighted in red) within reported statements and the summary.}
    \label{tab:sum_ex_2}
    \vspace{-0.5em}
\end{table}

\subsection{SRL Cue words}

The SRL system uses a pre-defined set of cues to identify matching verb predicates that signal attribution. The list of cues is provided in Figure \ref{tab:srl_cues}.



\begin{figure*}
    \footnotesize
    \begin{tabular}{|p{54em}|}
    \hline
accuse, affirm, allege, announce, argue, assert, aver, avouch, avow, blame, broadcast, claim, comment, confirm, contend, credit, declare, \\
\hline
declare, defend, describe, disclose, discuss, express, find, hint, imply, insinuate, insist, intimate, proclaim, profess, publish, reaffirm, \\
\hline
reassert, remark, repeat, report, say, state, tell, write, deny, gainsay, suppress, challenge, controvert, disagree, discount, discredit, dispute, \\
\hline
question, disavow, disclaim, protest, reject, repudiate, contradict, expect, add, think, believe, note, agree, plan, conclude, consider \\
\hline
    \end{tabular}
    \caption{Cues used by the Semantic Role Labeling (SRL) system to identify verb predicates corresponding to reported statements.}
    \label{tab:srl_cues}
\end{figure*}

\subsection{Coverage of Reported Statements in the Generated Summary}

Table \ref{tab:coverage_example_2} shows another qualitative example for our observation in Section 5.3 of the main paper that human summaries cover considerably more of the input reported statements compared to BART. We can see that the human summary covers information from statements 1,3,4,5 and 7 whereas the BART summary only covers statements 2 and 5. 

\subsection{Human Evaluation Guidelines}

        
        
\begin{figure*}[b]
    \centering
    \includegraphics[width=1.0\textwidth]{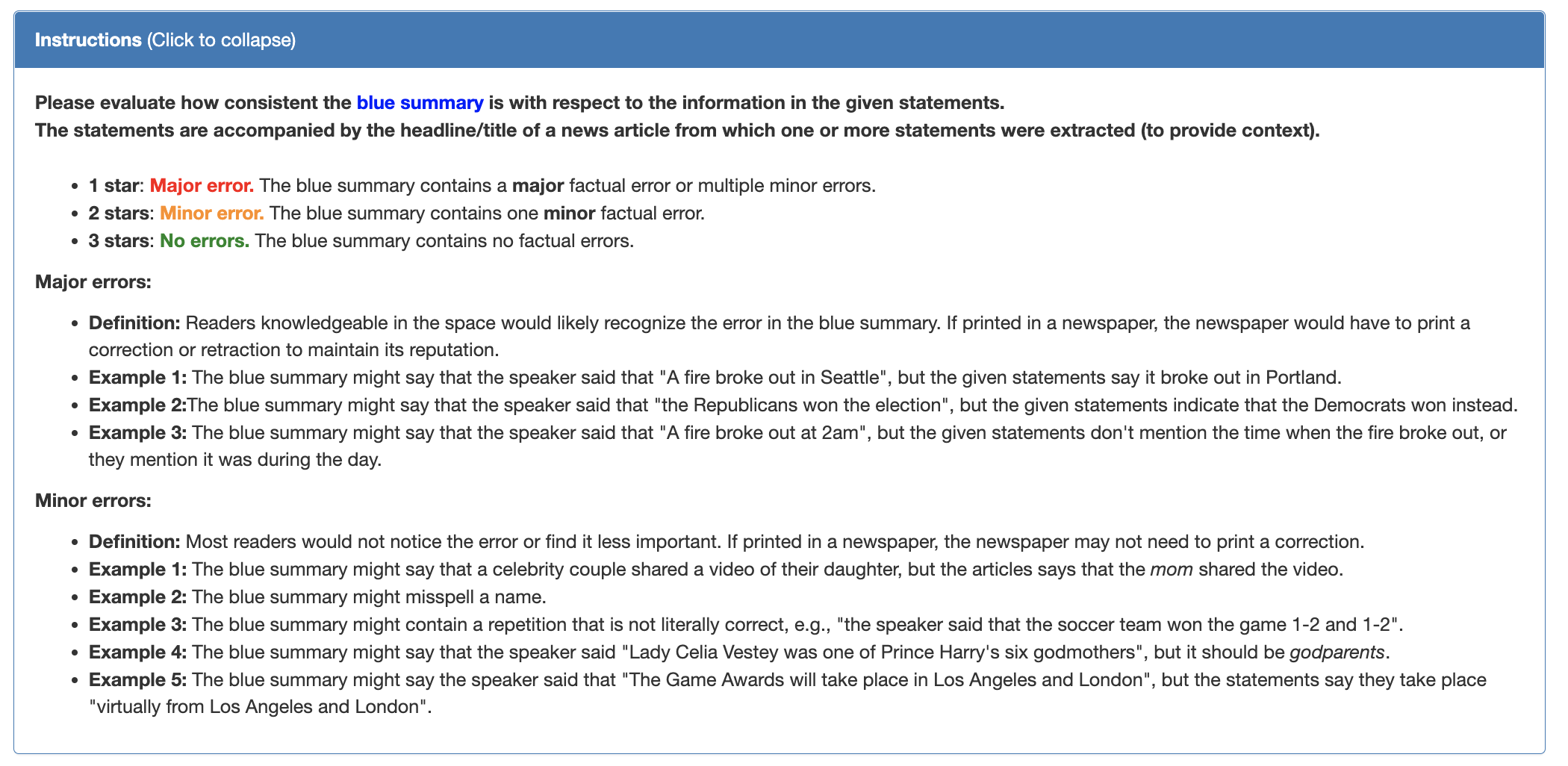}
    \caption{Guidelines for human evaluation of factual consistency of the generated summaries.}
    \label{fig:human_eval_consistency}
\end{figure*}
\begin{figure*}[b]
    \centering
    \includegraphics[width=1.0\textwidth]{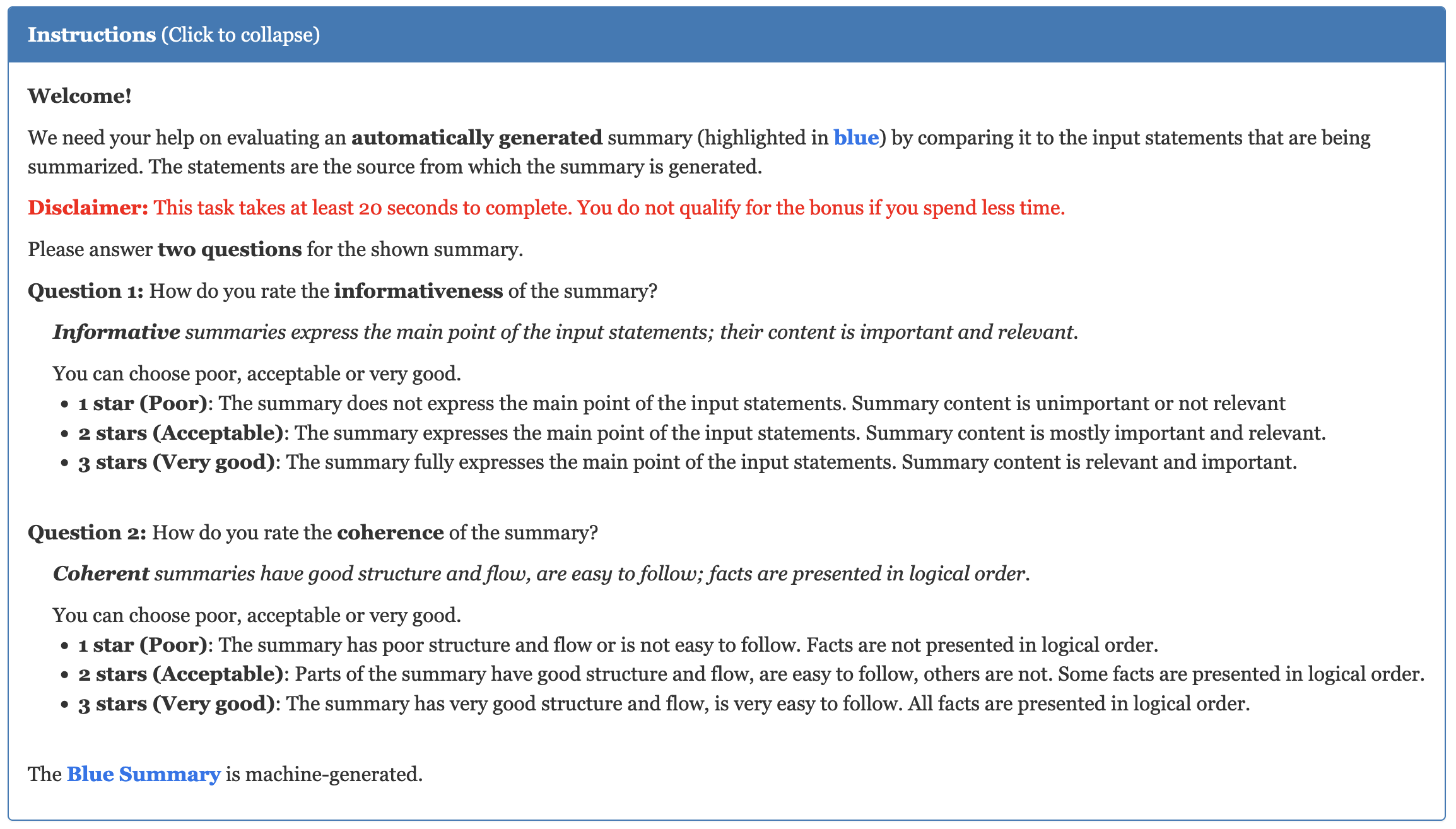}
    \caption{Guidelines for human evaluation of informativeness and coherence of the generated summaries.}
    \label{fig:human_eval_info_coherency}
\end{figure*}

Figure \ref{fig:human_eval_consistency} shows the human study guidelines for evaluation of factual consistency, and figure \ref{fig:human_eval_info_coherency} shows the guidelines for evaluating informativeness and coherence of the summaries. 

\end{document}